\documentclass[runningheads,a4paper]{llncs}

\usepackage{amssymb}
\usepackage{graphicx}
\usepackage{graphicx,subfigure}
\usepackage{multirow}

\usepackage{url}

\urldef{\mailsc}\path|deo.ratneel, c.rohitash}@gmail.com|

\begin{document}

\mainmatter  

\title{Stacked transfer learning  for tropical cyclone intensity 
prediction}

\titlerunning{Stacked transfer learning  for tropical cyclone intensity 
prediction}

%
%
\author{Ratneel Deo \and Rohitash Chandra \and 
Anuraganand Sharma }
\authorrunning{Ratneel Deo, Rohitash Chandra, Anuraganand Sharma}

\institute{School of Computing Information and Mathematical Science  
\\University
of the South Pacific, Suva, Fiji \\
\url{http://www.usp.ac.fj} \and Centre for Translational Data Science \\ 
The University of Sydney, Sydney NSW, Australia \url{ 
http://sydney.edu.au/data-science/}}

%
%

\maketitle

\begin{abstract}
Tropical cyclone  wind-intensity prediction is a challenging 
task  considering drastic changes  climate patterns over the last few decades. 
In order to develop robust 
prediction models, one needs to consider different characteristics of cyclones 
in terms of spatial and temporal characteristics. 
Transfer learning incorporates knowledge from a related  source dataset  to 
compliment a target datasets especially in cases where there is lack 
or data. Stacking is a form of ensemble learning focused for improving 
generalization that has  been recently  used for transfer learning 
problems which is referred to as transfer stacking. In this paper, we employ 
transfer stacking as a means of studying the effects of cyclones  whereby we 
evaluate if cyclones in different geographic locations can be helpful for 
improving generalization performance. Moreover, we use conventional neural 
networks for evaluating the effects of duration on cyclones in prediction 
performance.   Therefore, we develop an effective strategy that evaluates the 
relationships between different types of cyclones  
through transfer learning and conventional learning  methods via neural 
networks.

\end{abstract}

\section{Introduction}



 
Ensemble learning generally considers combination of  multiple 
standalone learning methods   to improve  generalization 
performance when compared to standalone approaches \cite{wolpert1992stacked}. 
Ensemble learning has been implemented with 
group of  neural networks that  are  trained as an ensemble with different 
configuration in parameter settings or initializations  for executing the same 
task\cite {ueda2000optimal,hansen1990neural}. Popular methods in ensemble 
learning involve stacking, bagging and boosting \cite{bauer1999empirical,drucker1997improving,wang2011comparative} . Stacked generalization is a form of ensemble 
learning  where  the predictions are combined  from the other ensembles.  The principal idea in stacked generalization is that more  learners would improve performance due the additional computational power.  \cite{wolpert1992stacked}. In the 
literature,  logistic regression has been commonly used as the combiner layer  
for stacking \cite{breiman1996stacked}. Stacking has been successfully used with 
both supervised and unsupervised learning 
\cite{smyth1999linearly,breiman1996stacked}. Recently, the approach has been 
applied to modifying emission kinetics of colloidal semiconductor nanoplatelets\cite{erdem2016temperature}.  


 Transfer learning   utilizes knowledge learned previously from related 
problems  into  learning models  in order  to have  faster training or  
better  or generalization performance \cite{pan2010survey}. Transfer learning 
incorporates knowledge from a related  problems (also known as source)  to 
compliment a target problem  especially in cases where there is lack 
of data or in cases where there is  requirement to speed up the learning 
process.   The approach has seen widespread applications with challenges  on the 
 type of knowledge  should be transferred in order to avoid negative transfers 
whereby the knowledge deteriorates the performance 
\cite{pardoe2010boosting,raina2007self,lawrence2004learning}.  Transfer Learning has recently been used for visual tracking, computer-aided detection \cite{gao2014transfer,shin2016deep} .  Transfer  
learning has been implemented  with  ensemble  learning  methods such as 
boosting and stacking
\cite{pardoe2010boosting}.  The approach in the case of transfer stacking is
implementing multiple learners previously trained on the source data in order to
have a single base learner. Simple stacking on the other hand uses 
multiple base learners\cite{pardoe2010boosting}. 


 


Tropical cyclone  wind-intensity prediction is a challenging 
task  considering drastic changes  climate patterns over the last few 
decades \cite{emanuel2005increasing,mu2009method,bennett1993tropical} . Previously, statistical models have been used  to forecast the 
movement and intensity of the cyclones.  Climatology and 
Persistence (CLIPER) was one of the earlier computer-based forecasting models which was 
able to give up to 5 days prediction, i.e., 72 hours of cyclone intensity 
\cite{Knaff2003}. Statistical hurricane intensity prediction scheme (SHIPS) was 
also used 
for cyclone intensity forecasts however, it was  restricted to storms over the ocean 
only \cite{DeMariaKaplan1999}.     There has been a growing interest in 
computational intelligence and machine learning methods  for cyclone 
prediction \cite{JinEtal2008,rajasekaran2008support,pham2016comparative}.    In the past, cyclone 
wind-intensity 
\cite{ChandraDayalCEC2015}  been tackled by cooperative  neuro-evolution of 
recurrent neural networks\cite{deo2016identification,chandra2016architecture}. 

In order to develop robust 
prediction models, one needs to consider different characteristics of cyclones 
in terms of spatial and temporal characteristics. Transfer learning can be used 
as a strategy to evaluate the relationship of cyclones from different 
geographic regions. Note that a negative transfer is considered when the source 
knowledge when utilized with the target data  contributes to poor 
generalization. This can be helpful in evaluating if the cyclones from a 
particular region is helpful in guiding as source knowledge for decision making 
by models  in other geographic locations. Moreover, it is also important to 
evaluate the effect of the  duration of a cyclone on the generalization 
performance given by the model. 

In this paper, we employ 
transfer stacking as a means of studying the effects of cyclones  whereby we 
evaluate if cyclones in different geographic locations can be helpful for 
improving generalization performs. We select the cyclones 
from last few decades  in the South Pacific and South Indian Oceans.  Firstly, 
we evaluate the performance of the standard neural networks when trained with 
different regions of the dataset. We then evaluate the effects on  
duration of the  cyclones in the South Pacific region and their contribution 
towards the neural networks generalization performance. Finally, we use 
transfer staking via ensembles and  consider the South Pacific region as the 
target model. We use South Indian Ocean as the source data and evaluate its 
impact on the South Pacific Ocean.   The backpropagation neural network  is  
used as  the stacked ensembles in the   transfer  stacking  method.

The rest of the paper is organized as follows.  Section \ref{method} presents 
the overview of transfer  stacking method and outlines the experimental setup.  
We discuss our experimental results in Section \ref{resul_disc} and conclude 
with directions for future research in  Section \ref{conclusion}.

\section{Methodology } 
\label{method} 

\subsection{Neural networks for time series prediction}
Time series are data of a series of events observed over a certain time period.  
In order to use neural networks for time series prediction, the original time 
series is reconstructed into a state-space vector with embedding dimension (D) 
and time-lag (T) through Taken's theorem \cite{Takens1981} . We consider the backpropagation algorithm which employs gradient descent  for training \cite{nevel1976stochastic}.  A feedforward neural network is defined 
by:

\begin{equation}
E(y_t)   =   g \bigg(  \delta_o + 
\sum_{h=1}^{H} v_{j} g \bigg(  \delta_h + \sum_{d=1}^{D} w_{dh} 
y_{t-d} )\bigg) 
\label{expected_y}
\end{equation}

where $\delta_o$ and $\delta_h$  are the bias weights for the output $o$ and 
hidden $h$ layer, respectively.  $V_j$ is the weight which maps the hidden 
layer $h$ to the output layer. $w_{dh}$ is the weight which maps $y_{t-d}$ to 
the hidden layer $h$ and $g$ is the activation function, which we assume to be a 
sigmoid function 
 for the hidden and output layer units. 
 
The root mean squared error which is generally used to test the performance of the FNN in given in equation \ref{rmse}.
 \begin{equation} 
RMSE =  \sqrt{\frac{ 1}{N }   \sum_{i=1}^{N} (y_i - \hat{y}_i)^2}
\label{rmse}  
\end{equation} 

\noindent where $y_i$ and $ \hat{y}_i$  are the observed  and 
predicted data, respectively. $N$ is the length of the observed data.

\subsection{Stacked transfer learning}
\label{transfer_learn}

\begin{figure}
\centering
    \hspace{-10mm}
    \includegraphics[width=70mm,height=50mm, 
    angle=0]{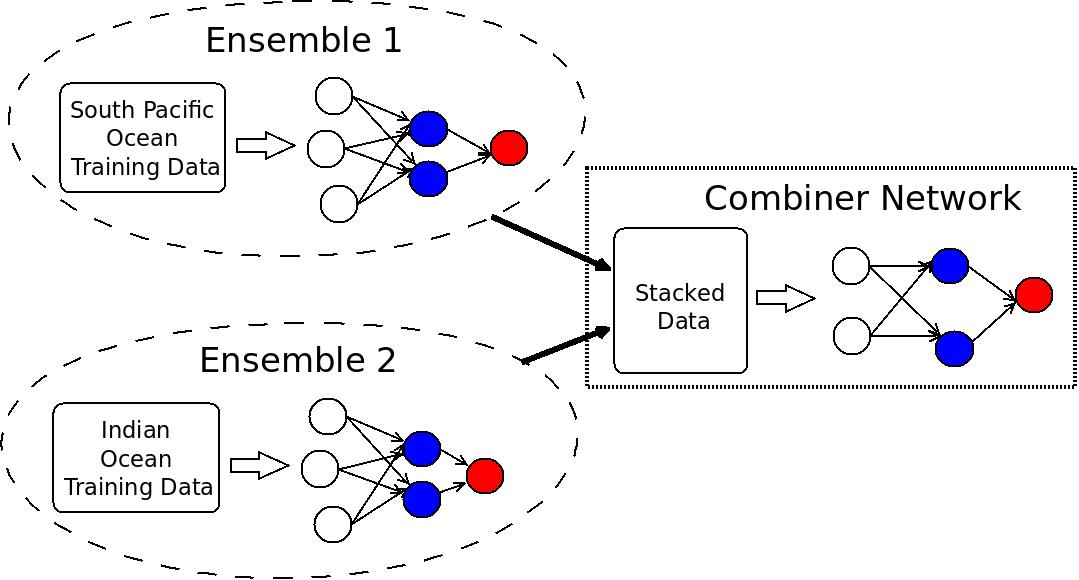}
    \caption{Neural Network Ensemble Model}
    \label{ensemble}
\end{figure}

Transfer learning is implemented via stacked ensembles which form source and 
target ensemble and a combiner ensemble that are  implemented using feedforward 
neural networks (FNNs). Enemble 1,2 and the combiner network in figure \ref{ensemble} show the FNNs . We 
refer to the transfer learning model  shown in figure \ref{ensemble} as transfer stacking hereafter.     Transfer stacking is implemented 
in two phases; phase one involves training individual ensembles from the 
network.  The second phase of the method trains a secondary prediction model 
which learns from the knowledge of the trained ensembles in phase 1.   Figure \ref{ensemble} shows the broader view of the ensemble 
stacking method where we have two ensemble models (FNNs) feeding knowledge into a secondary combiner network.  
The source and target ensembles  are  implemented using  FNNs with the same 
topology.  The combiner ensemble topology is depended on the number of 
ensembles used as  the source and target ensembles. 
Ensemble 1 considers  the 
South Pacific ocean  data while  Ensemble 2 considers the South  Indian ocean 
training data.   The datasets are described in section \ref{data_prc}.

The combiner ensemble is a feedforward  network that is trained on the  
knowledge processed by the ensembles. Backpropagation is  used for training the 
combiner network and the respective ensembles.  The processed knowledge comes 
from training data of the source and target data.  Knowledge was gathered from creating a stacked dataset 
that is a direct mapping from the training data.  This was achieved by 
concatenating the output of all the ensembles into a new stacked data file. The  stacked dataset encompassing the knowledge of the ensembles  is then used to train the combiner FNN network.

Similar to the two step training process, the testing done in two phases. The testing data was also passed through the stacking module to generate a stacked testing data.  The stacked testing data was then used in the combiner network to measure the generalization performance of transfer stacking .   


\subsection{Data Processing}
\label{data_prc}
The South Pacific ocean cyclone wind intensity data and South Indian Ocean 
cyclone wind 
intensity data for the year 1985 to 2013 had been used for these experiments\cite{jtwc}.  
We divided the data into training and testing sets.  Cyclones occurring in 
the years 1985 to 2005 were used for training while the remaining data was used 
to test for generalization performance. The consecutive  cyclones in the 
training and testing set was concatenated into a time series for effective 
modeling. The data was normalized to be in the range [ 0, 1].

\section{Experiments and Results}
\label{resul_disc}

In this section, we  present the experiments that we had used to test the transfer stacking for cyclone intensity prediction.  We later present our results based on the root mean squared error.

\subsection{Experiment Design}
\label{exp-setup}
 Standalone feed forward neural networks that were trained using the same back 
propagation learning algorithm used in the experiments.  Stochastic gradient 
descent was used with 2000 epochs training time.  30 experimental runs of each 
of the experiments was done in order to provide mean performance. The 
experiments were designed as presented in the list.

 \begin{enumerate}
  \item Experiment 1: Vanilla FNN trained on all South Pacific ocean training 
data and tested on South Pacific Ocean testing data.
  
  \item Experiment 2: Vanilla FNN trained on all Indian ocean training data and 
tested on South Pacific Ocean testing data.
  
  \item Experiment 3: 6 experiments with vanilla FNNs trained on subset of South 
Pacific ocean training data and tested twice. Once with the entire South Pacific 
Ocean testing data followed by the testing with a subset of the testing data . 
The subsets were created by grouping cyclones of similar lengths into classes. 
Each class of cyclones had been trained and tested with vanilla FNN model.  We 
formulated the following experiments with vanilla FNNs;
  
  \begin{enumerate}

      \item  $[0-3]$ day old cyclones in training set and tested with full 
testing set as well as $[0-3$ day old cyclones in testing set.
      \item  $[3 - 5]$ day old cyclones in training set and tested with full 
testing set as well as  $[3 - 5]$ day old cyclones in testing set.
      \item  $[5 - 7]$ day old cyclones in training set and tested with full 
testing set as well as  $[5 - 7]$ day old cyclones in testing set.
      \item  $[7 - 9]$ day old cyclones in training set and tested with full 
testing set as well as  $[7 - 9]$ day old cyclones in testing set.
      \item  $[9 - 12]$ day old cyclones in training set and tested with full 
testing set as well as  $[9 - 12]$ day old cyclones in testing set.
      \item  cyclones older than 12 days in training set and tested with full 
testing set as well as  cyclones older than 12 day in testing set.
   \end{enumerate}
  
  \item Experiment 4: Transfer learning based stacked ensemble method for 
predicting South Pacific ocean tropical cyclone intensity.  The mechanics of 
this method is given in section \ref{transfer_learn}. 

\item Experiment 5:  Two separate experiments were done in this experiment;
    \begin{enumerate}
    \item  FNN trained on 1985 - 1995 cyclones from south pacific data and 
tested on South Pacific Ocean testing data.
    \item  FNN trained on 1995 - 2005 cyclones from south pacific data and 
tested on South Pacific Ocean testing data.
    \end{enumerate}

 \end{enumerate}


\subsection{Results}

We present the results of the prediction of  tropical cyclones intensity in 
the South Pacific form the year 2006 to 2013.  The root mean squared error (RMSE) was  
 used to evaluate the performance as given in equation \ref{rmse}.

\begin{table}
 \centering
 \caption{Generalization performance   }
 \begin{tabular}{cc}
 \hline
  Experiment  & RMSE  \\
 \hline
  1 & $0.02863 \pm 0.00042$ \\
  2  & $0.03396 \pm 0.00075$ \\
  4 & $0.02802 \pm 0.00039$ \\
  
  \hline

 \end{tabular}
\label{tab:res}

\end{table}

Table \ref{tab:res} gives the generalization performance of the respective  
methods  on the testing data.  The results show that all the models have 
similar performance.  It looks at results from experiments 1, 2 and 4.


\begin{table}
 \centering
 \caption{Experiment 3: Performance of  FNN on different categories of cyclones}
 \begin{tabular}{cccc}
 \hline
  Cyclone Category  & Training RMSE &  Categorical Testing RMSE & Generalization RMSE\\
 \hline
  0-3 day  & $0.03932 \pm 0.00173$  & $0.05940 \pm 0.00329$ & $ 0.20569 \pm 0.01339$\\
  3-5 day  & $0.03135 \pm 0.00050$  & $0.02580 \pm 0.00044$ & $ 0.05200 \pm 0.00532$\\
  5-7 day  & $0.03265 \pm 0.00027$  & $0.02504 \pm 0.00025$ & $ 0.03070 \pm 0.00172$\\
  7-9 day  & $0.02831\pm 0.00033$  & $0.03799 \pm 0.00065$ & $ 0.03207 \pm 0.00055$\\
  9-12 day  & $0.03081\pm 0.00025$  & $0.02700 \pm 0.00059$ & $ 0.03154 \pm 0.00047$\\
  $\pm$ 12  day  & $0.02819\pm 0.00019$  & $0.03579 \pm 0.00037$ & $ 0.02875 \pm 0.00025$\\
  
  \hline

 \end{tabular}
\label{tab:period}

\end{table}

Table \ref{tab:period} gives the performance of various categories of data on the two testing datasets; category based testing and generalized testing. Category based testing is done on cyclones that belong that particular category of the testing data.  Generalization testing was done on the entire testing dataset.  The results show that cyclones that ended in under three days were not good predictors for the generalization data as they had higher testing error.  Similarly, cyclones that had duration of 3-5 days had poor performance as well giving a larger error however not as large as is predecessor.

Cyclones with duration of 5-12 days had very similar generalization performance.  These categories  of cyclones performed better than the smaller length cyclones by giving better prediction accuracy in terms of RMSE.  The final category of cyclones gave the best generalization performance.  Longer length cyclones, 12 days and over was matching the prediction accuracy of the best models that is, the standalone FNN model of control experiment 1, periodical and spatial analysis ensemble models. 

\begin{table}
 \centering
 \caption{Experiment 5: Performance of Vanilla FNN on independent decade training data }
 \begin{tabular}{ccc}
 \hline
  Vanilla FNN  & Training RMSE & Testing RMSE  \\
 \hline
  1985 - 1995 cyclones  & $0.04434 \pm 0.00073$  & $ 0.04671 \pm 0.00209$\\
  1995 - 2005 cyclones  & $0.03635 \pm 0.00062 $  & $ 0.05949 \pm 0.00201$\\

  \hline

 \end{tabular}
\label{tab:dec}

\end{table}

  Table \ref{tab:dec} shows the generalization performance achieved by the standalone FNN model trained with the decade 1 and decade 2 training data.  The generalization performance is rather poor with both the decades of data used independently when compared to the concatenation of all the training data as seen in  experiment 1.

\subsection{Discussion}

The experiments revealed interesting results about the respective
experiments. The first two experiments considered conventional
learning through neural networks (FNN)   for predicting tropical
cyclone wind intensity.   Experiments 1 considered cyclones where the
training and testing dataset from the same region (South Pacific
ocean) was used.  Experiment 2 considered cyclones where training data
from South Indian ocean and testing data from South Pacific ocean was
used.  According to the results, there was minimal difference in the
generalization performance in the South Pacific ocean, although the
training datasets considered different regions.  This implies that the
cyclones in the South Indian ocean have similar characteristics in
terms of the change of wind-intensity when compared to South Pacific
ocean.  Note that the size of the South  Indian ocean dataset was
about three times larger when compared to South Pacific ocean.

Furthermore, Experiment 3 featured an investigation into the effects
of the duration of cyclones  (cyclone lifetime) on the generalization
performance. This was done only for the case of the South Pacific
ocean.  Note that the generalization performance is based on the test
dataset that includes all the different types of cyclones that
occurred between year 2006 to year 2013. As seen in the results,
cyclones with shorter durations were not effective when considering
the generalization performance.  It seems that the shorter duration of
the cyclones did not give enough information to feature essential
knowledge for the longer duration of cyclones in the test dataset.
The category with  the cyclones with longest lengths gave the best
performance in terms of generalization performance .This implied that it had known about all the phases of the life cycle
of the cyclone, thus it was able to effectively predict all classes of
cyclones.


Transfer stacking via neural network ensembles was done in Experiment
4 where training dataset from South  Indian ocean was used as source
datasets and cyclones in the South Pacific Ocean was used in the
target datasets.  The generalization performance here was similar to
conventional neural networks given  Experiment 1 and 2.  This shows
that   the source data (South Indian ocean) did not make significant
contributions towards improving the generalization performance,
however, we note that there was not a negative transfer of knowledge
as the performance was not deteriorated. Therefore, the knowledge from
the cyclone behavior or change in wind-intensity  Indian ocean is
valid and applicable for South Pacific ocean. Further validation can
be done in future through examining the track information of the
cyclones from the respective regions. This will add further insights
of transfer learning through stacking and establish better knowledge
about the relationship of the cyclones in the respective regions.

\section{Conclusion}
\label{conclusion}
The paper presented  transfer stacking as a means of studying the contribution of cyclones in different geographic locations for improving generalization performance in predicting tropical cyclone wind intensity for the South Pacific ocean. We then evaluated  the effects on   duration of the  cyclones in the South Pacific region and their contribution towards the neural networks generalization performance.  Cyclone duration was seen to be a major contributor in the prediction of cyclone intensity. 

We found that cyclones with duration 
of over 12 days could be used as good representative for training the neural 
networks with competitive prediction intensity. Furthermore, the results 
show that the Indian Ocean source  dataset does not significantly improve the 
generalization performance of the   South Pacific target problem in terms of 
generalization performance. The contributions of the Indian ocean data was negligible as the knowledge about cyclone intensity prediction was sufficiently learned from the South Pacific data.  The change in geographical location was unable to provide any new knowledge which would improve generalization. 

Further work can review other cyclone regions into the transfer learning 
methodology to further improve the generalization performance. The approach can 
be extended to prediction of cyclone tracks in the related regions. Recurrent neural networks (RNN) could be used as ensemble learners to identify any temporal sequences that FNN was unable to learn as RNNs have shown better modeling performance.  Hybrid ensemble models of neural networks could also  be developed that use evolutionary algorithms for training the FNN together with backpropagation to improve the modeling.


\bibliographystyle{splncs03}
\bibliography{reference,references}

\end{document}